\documentclass[runningheads]{llncs}
\usepackage[T1]{fontenc}
\usepackage{threeparttable}
\usepackage{xcolor}
\usepackage[hidelinks]{hyperref}
\usepackage{flafter}
\usepackage{placeins}
\usepackage{graphicx}
\usepackage{amsmath}
\usepackage{multirow}
\usepackage{booktabs}
\usepackage{tikz}
\usetikzlibrary{positioning}
\begin{document}
%

% \title{Galop: Scaling Genetic Programming to Billions of Individuals}
\title{Galop: Billion-Scale Genetic Programming on a Single GPU}

% %
\author{Xavier Hill Roy\inst{1}\orcidID{0009-0009-1600-3525} \and
Stephen Kelly\inst{1}\orcidID{0000-0002-6071-4705}}
% %

\institute{McMaster University, Hamilton ON  L8S 1C7, Canada}
\maketitle              % typeset the header of the contribution

\begin{abstract}
Genetic programming (GP) evolves populations of heterogeneous programs, whereas GPUs achieve their highest throughput when groups of threads execute the same instructions. To better align GP evaluation with GPU execution, we present Galop, a hybrid CPU–GPU linear GP system that assigns one individual to each CUDA warp, a group of 32 threads that execute the same instructions across different fitness cases. We compare Galop against EvoGP, a recent high-performance GPU-based tree GP system, on data-rich symbolic regression across two unique GPUs, varying population size and fitness-case count. Galop completes evolutionary runs faster than EvoGP in 27 of the 28 main comparisons, with speedups of up to $13.6\times$. Solution quality comparisons favour EvoGP overall, although Galop remains competitive on several problems. Galop also completes a full generation with 2.15 billion individuals on a single H100 GPU, approximately $4.0\times$ the largest population achieved by any other tested GP system. All five runs at this population complete 100 generations, with a median wall-clock time of 2.23 hours. These results make it practical to study GP populations in the billions on a single GPU. Galop's source code, experimental data, and
reproducibility scripts are publicly available
at \url{https://github.com/xavierhillroy/Galop}.

\end{abstract}

\section{Introduction}
\label{sec:introduction}

Large populations are of growing interest across evolutionary computation. In digital evolution, which studies evolving populations of computational organisms, population scale has been linked to diversity and niche formation, and proposed as a potential precondition for open-ended evolution~\cite{dolson_digital_2021,taylor_open-ended_2016}. In evolutionary approaches to machine learning, recent work also suggests that larger populations can improve evolutionary search~\cite{yu_scaling_2026}. However, larger populations require evaluating more individuals every generation, making such experiments computationally expensive. GPUs offer substantial parallel throughput, and modern GPU-based genetic programming (GP) systems can outperform CPU implementations by orders of magnitude on sufficiently large workloads~\cite{wu_enabling_2026,de_vries_kozax_2025}. The challenge is how to use this hardware efficiently when evaluating populations of different programs.

\begin{figure}[ht]
\centering
\includegraphics[width=0.7\columnwidth,page=7]{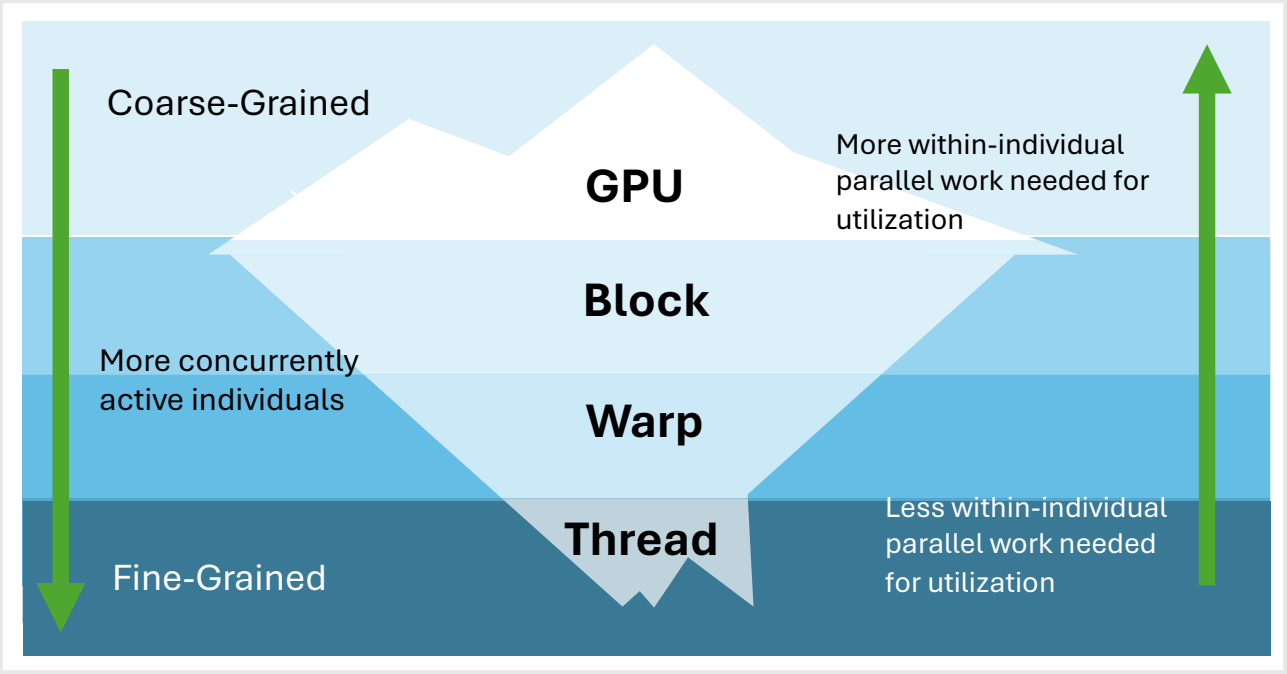}
\caption{
\textbf{The Execution Granularity Iceberg.}
Finer-grained execution increases the number of concurrently active
individuals, while coarser mappings rely on internal parallel work
for utilization. This is a modified version of the execution
granularity iceberg from our previous work~\cite{hill_roy_search_2026}.
}
\label{fig:iceberg}
\end{figure}

GP individuals can differ in the instructions they execute, whereas GPUs are most efficient when groups of threads execute the same instructions. In CUDA (Compute Unified Device Architecture), a kernel is a function executed on the GPU across a grid of thread blocks. Each block contains threads organized into warps of 32, which execute instructions together \cite{noauthor_cuda_nodate}.
% Threads within awarp execute together, so different instruction paths must
% be executed sequentially~\cite{nickolls_gpu_2010}.
These execution units form a hierarchy from device to block to warp to thread (Fig.~\ref{fig:iceberg}).

GPU-based GP can exploit two sources of parallelism: data-level parallelism, where an individual is evaluated across independent fitness cases, and population-level parallelism, where different individuals are evaluated concurrently. We denote population size by $P$ and the number of fitness cases by $D$.

In symbolic regression, for example, a program can be evaluated on many input--output examples independently. When $D$ is large, data-level parallelism can keep a GPU highly utilized even when only a few individuals execute concurrently. In agent-based systems, including reinforcement learning and robot control, an agent interacts with its environment over a sequence of timesteps. Each action influences the next state, so successive timesteps cannot be evaluated independently. When only a few independent simulations are available per individual, this limits data-level parallelism, making population-level parallelism more important. We previously identified these workloads as a motivation for exploiting the population axis~\cite{hill_roy_search_2026}.

The amount of population parallelism exposed depends on the GPU execution unit assigned to each individual (Fig.~\ref{fig:iceberg}). Prior systems have explored mappings ranging from the entire device to individual threads~\cite{zhang_speeding_2022,robilliard_genetic_2009}. Coarser mappings, such as device- or block-level execution, assign more threads to each individual. This allows them to exploit substantial data-level parallelism when $D$ is large, but limits the number of different individuals executing concurrently.

Finer mappings, such as warp- or thread-level execution, expose more population-level parallelism. However, assigning different programs to threads within the same warp can cause warp divergence, reducing GPU utilization~\cite{nickolls_gpu_2010}. Mapping one individual to each warp provides an intermediate approach: the threads within a warp execute the same program across different fitness cases, while separate warps evaluate different individuals. This suggests a workload-dependent trade-off: coarser mappings can be effective for data-rich problems such as symbolic regression, where each individual provides substantial parallel work, while warp-level mappings may be better suited to large populations with more limited parallel work per individual.

We present Galop, a hybrid CPU--GPU linear GP (LGP) system~\cite{brameier_linear_2010} built around this warp-per-individual mapping. Selection and variation run on the CPU, while fitness evaluation runs on the GPU.

Although Galop is motivated by workloads with limited parallel work per individual, we evaluate it first using symbolic regression. This is a challenging setting for Galop: when $D$ is large, each individual provides abundant independent fitness-case work that coarser GPU mappings can exploit efficiently. Symbolic regression also provides a controlled setting in which population size $P$ and fitness-case count $D$ can be varied independently without changing the underlying regression problem. We can therefore evaluate Galop across the $P\times D$ design space, including regimes where coarser mappings are expected to be strong.

Our primary comparison is against EvoGP~\cite{wu_enabling_2026}, a high-performance GPU-native tree GP (TGP) system that adapts its execution granularity to the number of fitness cases. We compare solution quality using predefined evolutionary settings, without tuning either system to achieve its best possible performance on individual problems. Beagle, SRGPU, and Kozax provide additional computational performance comparisons~\cite{haut_gpu_2026,zhang_speeding_2022,de_vries_kozax_2025}.

Across two unique GPUs, Galop is faster than EvoGP in 27 of the 28 primary configurations, with speedups of up to $13.6\times$. Solution quality comparisons favour EvoGP overall, although Galop achieves significant wins on several problems. On a single H100, Galop also evolves 2.15 billion individuals for 100 generations in a median of 2.23\,h, a population approximately $4.0\times$ that of the next-largest tested GP system.

This paper makes four contributions: (1)~Galop, a complete and open-source hybrid CPU--GPU LGP system realizing warp-per-individual execution; (2)~an end-to-end speed characterization across population size and fitness-case count on consumer and datacenter GPUs; (3)~a matched-population quality comparison against EvoGP; and (4)~a single-GPU population-capacity study demonstrating sustained evolution beyond two billion individuals and examining the factors that determine the practical population limit.

\section{Related GPU-GP Systems}
\label{sec:device_block}

Rather than providing an exhaustive history of GPU-based GP, this section focuses on the contemporary systems used as comparisons in our experiments and on how they map individuals onto the GPU execution hierarchy (Fig.~\ref{fig:iceberg}). A broader historical treatment of GPU-GP execution is provided in our previous position paper \cite{hill_roy_search_2026}.

SRGPU performs selection and variation on the CPU, while fitness evaluation runs on the GPU~\cite{zhang_speeding_2022}. It assigns the entire GPU to one individual at a time, evaluating its fitness cases in parallel before moving to the next individual. SRGPU therefore relies primarily on data-level parallelism ($D$).

EvoGP is a TGP system that performs fitness evaluation, selection, and variation entirely on the GPU~\cite{wu_enabling_2026}. It adapts its execution granularity to the number of fitness cases $D$. For smaller $D$, it assigns one or more CUDA blocks to each individual, while for larger $D$, it assigns the entire device to one individual at a time. 

Beagle is an LGP system that assigns one CUDA block to each individual~\cite{haut_gpu_2026,basin_introduction_2026}. Selection and variation run on the CPU, while fitness evaluation runs on the GPU. Its implementation limits full-batch evaluation to $D\leq1024$ (the maximum number of threads per CUDA block), whereas Galop handles larger datasets by processing fitness cases in successive groups of 32.

Kozax is a tree GP system built using JAX, a framework that compiles array operations for GPU execution~\cite{de_vries_kozax_2025}. It represents programs as fixed-size arrays and evaluates multiple individuals and fitness cases through batched operations. Unlike the other systems, Kozax does not explicitly assign individuals to CUDA execution units such as blocks or warps.  Instead, JAX's compiler determines how these operations are mapped to GPU execution units during compilation.

Historically, Robilliard et al.\ also compared explicit thread-per-individual and block-per-individual mappings and discussed placing multiple programs within one block \cite{robilliard_genetic_2009}. Galop builds on this broader exploration of execution granularity by adopting the warp as the unit of individual evaluation.

\section{Galop}
\label{sec:Galop}
Galop is a hybrid CPU--GPU LGP system with compact, variable-length programs. Selection and variation run on the CPU for flexibility in handling irregular operations, while fitness evaluation runs on the GPU to exploit parallelism. In LGP, each individual is a sequence of instructions that operate on registers, rather than an expression tree.

Each LGP instruction occupies 32 bits and encodes an opcode, destination, and two operands. The implementation evaluated in this paper does not support data-dependent jumps or loops, so each program follows a fixed instruction sequence during evaluation. Supporting such control flow may introduce warp divergence across fitness cases and complicate the removal of non-effective code. Galop uses one-point crossover and three mutation operators: micro mutation modifies a field of an existing instruction, while insertion and deletion add or remove instructions, respectively. Galop caches fitness values for unchanged elites, avoiding redundant evaluation across generations.

\subsection{Warp-Per-Individual Evaluation}
\label{sec:Galop-warp-evaluation}

\begin{figure}[!htbp]
\centering
\includegraphics[width=0.85\columnwidth]{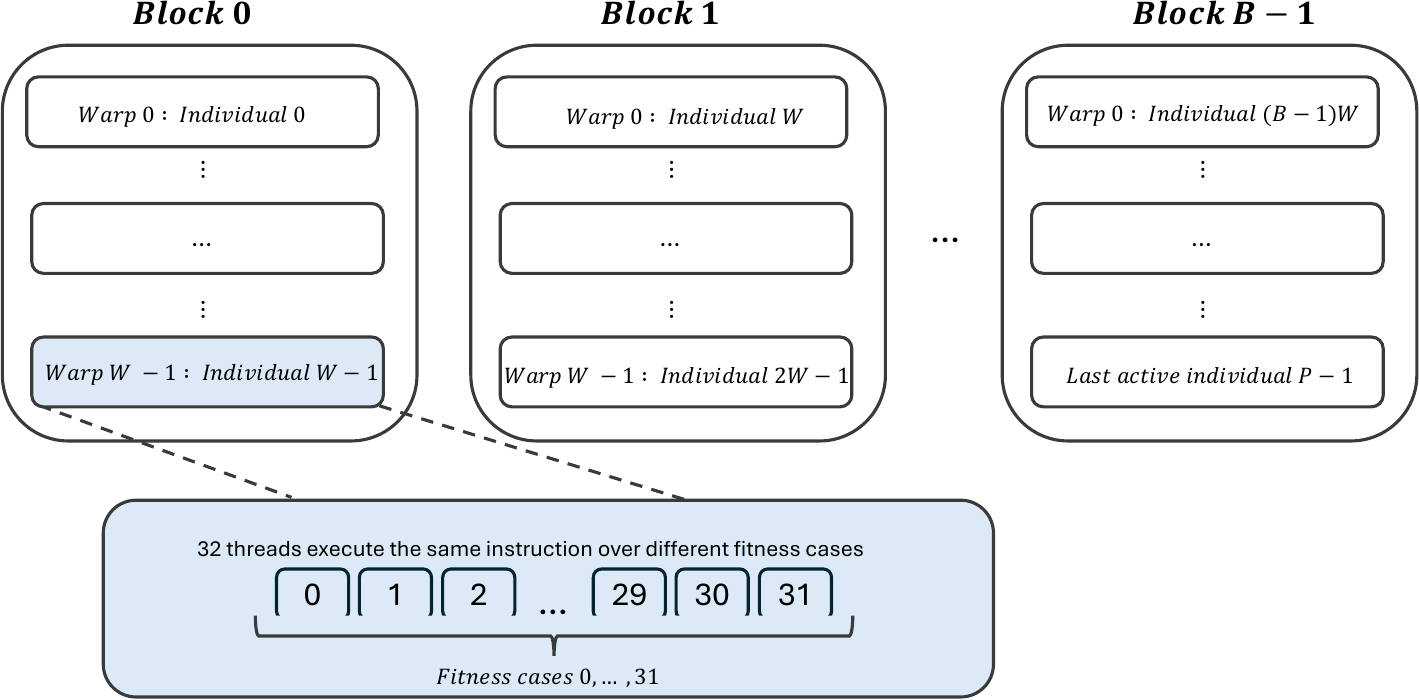}
\caption{
\textbf{Galop's warp-per-individual mapping.}
Each CUDA warp evaluates one individual, with its 32 threads executing the same program over different fitness cases. Each block contains $W$ warps, giving $B=\lceil P/W\rceil$ blocks (B) for a population of $P$ individuals.
}
\label{fig:Galop-warp-mapping}
\end{figure}

Galop assigns one individual to each CUDA warp. The 32 threads within a warp execute the same program across different fitness cases, while separate warps are assigned to different individuals (Fig.~\ref{fig:Galop-warp-mapping}).

For $D\le32$, one warp covers the complete fitness set. For larger datasets, the same warp processes successive groups of 32 cases and aggregates their errors to produce the individual's fitness.

In our experiments, each CUDA block contains eight warps (256 threads), with each warp assigned to a different individual. Grouping multiple warps into a block makes more efficient use of the GPU's limited resident-block slots, allowing more warps to execute concurrently on each SM.

\subsection{Instruction Packing}
\label{sec:instruction-packing}

On the CPU, Galop maintains programs in fixed-capacity buffers. Before GPU evaluation, their instructions are packed into a contiguous array, with an offset array identifying each program. This avoids storing or transferring unused padding on the GPU. The packed population is transferred once per generation, while fitness cases remain resident on the device. After evaluation, all fitness scores are transferred to the CPU as a single array for selection and variation.

LGP programs may contain introns, instructions that do not affect the program's output. In the optional intron-removed layout, Galop removes these instructions before transfer and evaluation~\cite{brameier_comparison_2001}, further reducing memory use and computation without changing program behaviour.

\section{Experimental Methodology}
\label{sec:methodology}
Our evaluation comprises three experiments: end-to-end evolutionary speed, solution quality at matched populations, and population capacity. This section describes the systems, matched experimental settings, datasets, measurements, and protocols.

\subsection{Systems, Configuration, and Hardware}
\label{sec:experimental-systems}

EvoGP is the primary baseline because it is a recent high-performance GPU-GP system and permits broad experimental alignment with Galop~\cite{wu_enabling_2026}. Table~\ref{tab:configuration} summarizes the experimental configuration. We match shared settings between Galop and EvoGP, including a common primitive set spanning arithmetic, trigonometric, and comparison operations.

\begin{table}[!htbp]
\centering
\begin{threeparttable}
\caption{Primary Galop--EvoGP experimental configuration.}
\label{tab:configuration}
\footnotesize
\setlength{\tabcolsep}{4.5pt}
\begin{tabular}{lll}
\toprule
Setting & Galop & EvoGP \\
\midrule

Representation & LGP & TGP \\

Primitives
& $+,-,\times,\div,\sin,\cos,<,>$
& $+,-,\times,\div,\sin,\cos,<,>$ \\

Constants & fixed pool\tnote{a} & fixed pool\tnote{a} \\
Objective & MSE & MSE \\
Selection & tournament 3 & tournament 3 \\
Elitism & 10\% & 10\% \\

Crossover
& one-point, $p=1.0$
& subtree, $p=1.0$ \\

Mutation
& micro / insertion / deletion
& subtree replacement \\

Mutation rate(s)
& $0.8\;/\;(0.8\times3)\tnote{b}\;/\;0.75$
& $0.8$ \\

Size limit
& 50 instructions
& 101 nodes \\

\bottomrule
\end{tabular}

\begin{tablenotes}[flushleft]
\footnotesize
\item[a] Both systems use the same pool of eight constants,
whose values remain fixed throughout evolution.

\item[b] Insertion mutation uses three independent trials,
each with probability $0.8$, subject to the
50-instruction limit.
\end{tablenotes}

\end{threeparttable}
\end{table}
Beagle, SRGPU, and Kozax provide additional reference systems for computational context \cite{haut_gpu_2026,zhang_speeding_2022,de_vries_kozax_2025}.

Experiments use an NVIDIA GeForce RTX 5070 Ti (16\,GB) and an NVIDIA H100 (80\,GB). The Galop--EvoGP speed experiments run on both GPUs; matched-population quality runs on the RTX 5070 Ti, while the additional reference-system timing and population-capacity experiments run on the H100. All Galop experiments use 16 OpenMP host threads.

\subsection{Problems and Datasets}
\label{sec:datasets}

Symbolic regression evolves expressions to approximate a target function from input--output examples, making the number of fitness cases $D$ a controllable source of parallel work per individual. Galop and EvoGP use identical training and test data and optimize training mean squared error (MSE). We report held-out test MSE for the best individual from each run. For exactly representable targets, a run is considered solved when test MSE falls below $10^{-12}$.

\paragraph{Nguyen.} Nguyen 1--6~\cite{uy_semantically-based_2011} are single-input polynomial and trigonometric benchmarks, all exactly representable using the shared primitive set. Each problem uses 100 training and 200 test cases sampled from $U[-1,1]$ with fixed seeds.

\paragraph{Feynman I.9.18.} Feynman I.9.18~\cite{udrescu_ai_2020} is a nine-input Newtonian-gravitation problem, also exactly representable using the shared primitive set. We sample 1,024 training and 2,048 test cases with fixed seeds.
\paragraph{California Housing.} California Housing~\cite{kelley_pace_sparse_1997} is an eight-input regression dataset with 20,640 observations. We use an 80/20 train/test split (16,512/4,128 cases) and standardize features and targets using training-set statistics only.

\subsection{Measurement, Statistics, and Verification}
\label{sec:measurement}

All timed experiments have exclusive use of the GPU and report median wall-clock time over five runs unless otherwise stated. EvoGP's detail logging is disabled during timing to avoid logging overhead.

End-to-end wall-clock time is our primary performance measure. GPops/s~\cite{langdon_many_2010} is useful within a common representation but is not directly comparable between TGP and LGP: TGP counts tree nodes, including operators and terminals, whereas LGP counts instructions containing an operator and its operands. Galop's 50-instruction cap and EvoGP's 101-node cap provide roughly comparable operator capacity, but do not make GPops/s directly comparable.

For quality comparisons, the independent evolutionary run is the unit of analysis. Best-of-run test-MSE distributions are compared within the predefined inference families using two-sided Mann--Whitney $U$ tests \cite{mann_test_1947} with Holm correction~\cite{holm_simple_1979} and Vargha--Delaney $A_{12}$~\cite{vargha_critique_2000}; $A_{12}<0.5$ favours Galop. Non-significant comparisons are reported as undecided rather than equivalent.

We validate Galop using unit tests, CPU reference tests, and CPU--GPU comparisons on random and evolved populations. We also verify that results are independent of host thread count and that packed and intron-removed programs produce identical fitness.

\subsection{Experimental Protocol }
\label{sec:experimental-protocols}
\paragraph{End-to-End Speed.} Complete-run performance is measured over $G=100$ generations on the Nguyen~2 symbolic-regression target. For each $D$, the same resampled dataset is supplied to every system, allowing the amount of fitness-case work to vary without changing the underlying problem. The primary grid is
\[
P\in\{10^{3},10^{4},10^{5},10^{6}\},
\qquad
D\in\{32,\,1024,\,32768\}.
\]
The 12 $P\times D$ combinations are evaluated on both GPUs, with $D=131{,}072$ additionally tested on the H100, giving 28 primary configurations in total. Further H100 boundary tests use $D\in\{262{,}144,524{,}288\}$ for $P\le10^{5}$. The reference systems are evaluated on the primary H100 configurations where supported.

Timing covers the complete evolutionary pipeline, not just GPU evaluation. After warm-up, five complete runs are timed for each system configuration and the median is reported. Timing spread is defined as $(t_{\max}-t_{\min})/t_{\mathrm{median}}$ over these five runs for each system. Comparisons with a spread exceeding $25\%$ in either system are considered unreliable. We remeasured the two affected Galop--EvoGP comparisons and excluded affected reference-system comparisons.

\paragraph{Quality at Matched Populations.} Head-to-head quality experiments use $G=100$ generations and $n=100$ independent runs per system and configuration, with both systems using the same seeds (0--99). Nguyen 1--6 are evaluated at $P\in\{10^{3},10^{4},10^{5}\}$, while California Housing and Feynman I.9.18 are evaluated at $P\in\{10^{4},10^{5}\}$, giving 22 primary problem--population combinations. The Nguyen suite forms one Holm family at each population size, while California and Feynman each form one family.

As a sensitivity analysis, we repeat the eight comparisons at $P=10^4$ with Galop unchanged and EvoGP using its published example settings ($n=100$, separate Holm family).

\paragraph{Population Capacity.} The third experiment measures each system's maximum population size on the H100. A population size is considered successful if the system completes one full generation on Nguyen~2 with $D=64$ within 1,200\,s (20 minutes), using one 80\,GB H100 and up to 1\,TB of host memory. The small fitness set emphasizes population size rather than fitness-case computation. Population sizes are increased until the first failure, with each size tested once. Galop is tested with both packed and intron-removed programs, as well as a stress test in which every program is initialized at the 50-instruction limit.

To test sustained evolution, we also perform complete 100-generation runs at $P=10^8$, $10^9$, and $2^{31}-2$ using Galop's intron-removed layout, with five runs at each population. Beagle is also tested at $P=10^8$ for comparison. We report median wall-clock time and the minimum--maximum range across the five runs. Galop's extreme-scale runs were conducted across multiple shared H100 nodes, while Beagle's runs were collected within one session. These measurements therefore include node-to-node variability and are not subject to the 25\% timing-spread criterion used for the end-to-end speed experiments.

\section{Results}
\label{sec:results}
This section characterizes Galop along three axes: end-to-end performance across the $P\times D$ design space, solution quality, and single-GPU population capacity.

\subsection{End-to-End Speed}
\label{sec:results-speed}
\begin{figure}[!htb]
\centering
\includegraphics[width=0.65\linewidth]{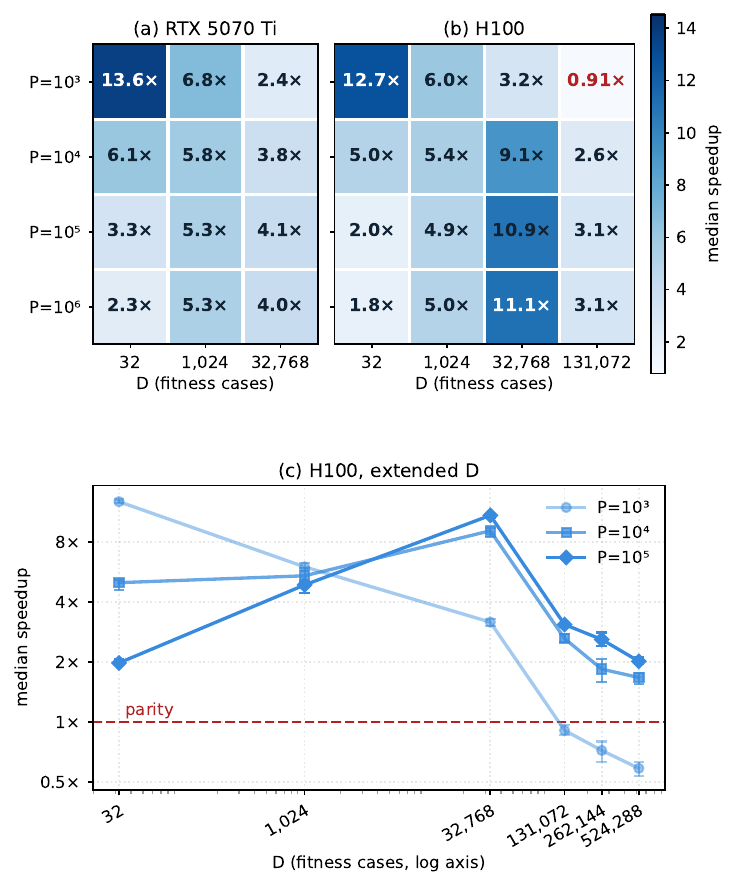}
\caption{
End-to-end Galop speedup over EvoGP on Nguyen~2, measured as EvoGP wall-clock divided by Galop wall-clock. (a,b) Primary population and fitness-case grid on the RTX~5070~Ti and H100. (c) H100 extension to larger fitness sets. Error bars show speedup bounds from the fastest and slowest observed runtimes over five runs. Values greater than one favour Galop.
}
\label{fig:whole-run-speedup}
\end{figure}

\begin{table}[!htb]
\centering
\caption{
Median end-to-end wall-clock times (s) for the
28 primary configurations over 100 generations on
Nguyen~2. Each value is the median of five timed
runs after warm-up. Bold indicates the sole configuration in which EvoGP
achieves a lower median runtime. Additional H100 boundary probes
are shown in Fig.~\ref{fig:whole-run-speedup}c.
}
\label{tab:whole-run-times}
\scriptsize
\setlength{\tabcolsep}{3pt}
\renewcommand{\arraystretch}{0.95}
\begin{tabular}{lllrrrr}
\toprule
GPU & $P$ & System & $D=32$ & $D=1{,}024$ &
$D=32{,}768$ & $D=131{,}072$ \\
\midrule

\multirow{8}{*}{RTX 5070 Ti}
& \multirow{2}{*}{$10^3$}
& Galop & 0.0086 & 0.0273 & 0.636 & -- \\
& & EvoGP & 0.117 & 0.187 & 1.55 & -- \\
\addlinespace[2pt]

& \multirow{2}{*}{$10^4$}
& Galop & 0.0319 & 0.153 & 3.93 & -- \\
& & EvoGP & 0.196 & 0.889 & 14.9 & -- \\
\addlinespace[2pt]

& \multirow{2}{*}{$10^5$}
& Galop & 0.244 & 1.38 & 36.6 & -- \\
& & EvoGP & 0.803 & 7.29 & 148 & -- \\
\addlinespace[2pt]

& \multirow{2}{*}{$10^6$}
& Galop & 3.31 & 14.3 & 371 & -- \\
& & EvoGP & 7.77 & 76.4 & 1,493 & -- \\

\midrule

\multirow{8}{*}{H100}
& \multirow{2}{*}{$10^3$}
& Galop & 0.0118 & 0.0317 & 0.691 & 2.66 \\
& & EvoGP & 0.150 & 0.190 & 2.19 & \textbf{2.41} \\
\addlinespace[2pt]

& \multirow{2}{*}{$10^4$}
& Galop & 0.0408 & 0.108 & 2.28 & 8.89 \\
& & EvoGP & 0.204 & 0.582 & 20.7 & 23.3 \\
\addlinespace[2pt]

& \multirow{2}{*}{$10^5$}
& Galop & 0.278 & 0.863 & 18.9 & 74.9 \\
& & EvoGP & 0.549 & 4.20 & 206 & 230 \\
\addlinespace[2pt]

& \multirow{2}{*}{$10^6$}
& Galop & 2.70 & 8.22 & 186 & 739 \\
& & EvoGP & 4.96 & 40.9 & 2,067 & 2,301 \\

\bottomrule
\end{tabular}
\end{table}

\begin{figure}[!htbp]
\centering
\includegraphics[width=0.75\linewidth]
    {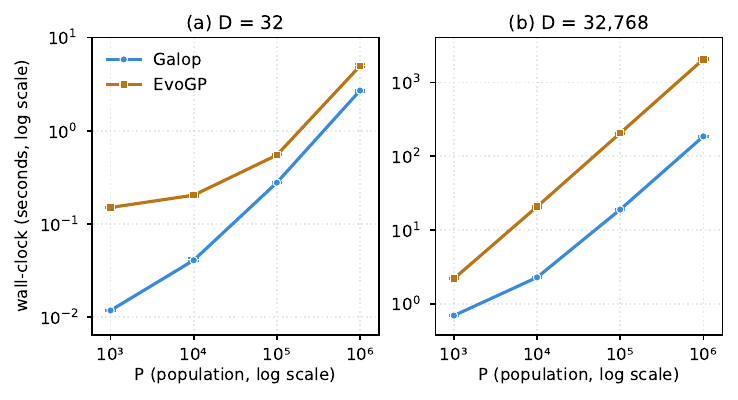}
\caption{
End-to-end runtime scaling with population size on the H100
for Nguyen~2 over 100 generations, at (a) $D=32$ and
(b) $D=32{,}768$. Points show median wall-clock time
over five runs, with minimum--maximum error bars.
Both axes are logarithmic. Error bars are smaller than
the markers in every configuration.
}

\label{fig:scaling}
\end{figure}

Galop completes the evolutionary run faster in 27 of 28 primary configurations across the two GPUs, with speedups of up to $13.6\times$ (Fig.~\ref{fig:whole-run-speedup}a,b). Table~\ref{tab:whole-run-times} reports absolute runtimes for the primary configurations. On the H100, the smallest configuration takes 12\,ms with Galop versus 150\,ms with EvoGP. At the largest configuration, $P=10^6$ and $D=131{,}072$, the respective runtimes are 739\,s and 2{,}301\,s.

At $D=32$, Galop's relative advantage decreases as population size grows, from $12.7\times$ at $P=10^3$ to $1.8\times$ at $P=10^6$ on the H100. However, its runtime continues to scale approximately linearly at larger populations: increasing $P$ from $10^5$ to $10^6$ increases runtime from 0.278\,s to 2.70\,s (Fig.~\ref{fig:scaling}a).

A timing breakdown helps explain this trend. At $D=32$, GPU kernel execution accounts for $36\%$ of Galop's runtime at $P=10^3$, falling to just $8\%$ at $P=10^6$. At the largest population, CPU selection and variation account for $39\%$ of runtime, while population packing and host--device transfers account for another $33\%$. Thus, when fitness evaluation is inexpensive, the surrounding evolutionary pipeline becomes the main computational cost.

At larger fitness sets, GPU evaluation dominates. On the H100, the kernel accounts for $71$--$76\%$ of Galop's runtime at $D=1{,}024$ and at least $96\%$ at $D=32{,}768$. For these fitness sets and $P\ge10^4$, Galop is approximately $5$--$11\times$ faster than EvoGP. Both systems approach linear runtime scaling at larger populations, while Galop maintains a substantial advantage (Fig.~\ref{fig:scaling}b).

At very large fitness sets and small populations, however, EvoGP becomes more competitive. At $P=10^3$, EvoGP overtakes Galop at $D=131{,}072$, and Galop's relative speedup falls to $0.58\times$ at $D=524{,}288$ (Fig.~\ref{fig:whole-run-speedup}c). At $P=10^4$ and $10^5$, Galop remains faster through the largest tested fitness set. These results show that the relative performance of the two systems depends on the balance between population-level and data-level parallelism.

The additional reference systems provide broader computational context. Across comparable H100 measurements, Galop is $1.2$--$1{,}019\times$ faster than SRGPU and $107$--${\sim}21{,}000\times$ faster than Kozax. Across the five Beagle comparisons satisfying the predefined $25\%$ timing-spread criterion, Galop is $10.4$--$20.7\times$ faster.

These speedups reflect Galop's combined execution mapping, compact representation, and CPU--GPU system design; they do not isolate the contribution of warp-level execution.  One design choice can be examined separately within Galop. On the RTX 5070 Ti at $P=10^5$ and $D=32$, grouping eight warps into each 256-thread block increases achieved GPU occupancy from $47.3\%$ with one warp per block to $82.2\%$, as measured using Nsight Compute.

\subsection{Quality at Matched Populations}
\label{sec:results-quality}

\begin{figure}[!htbp]
\centering
\includegraphics[width=\linewidth]
    {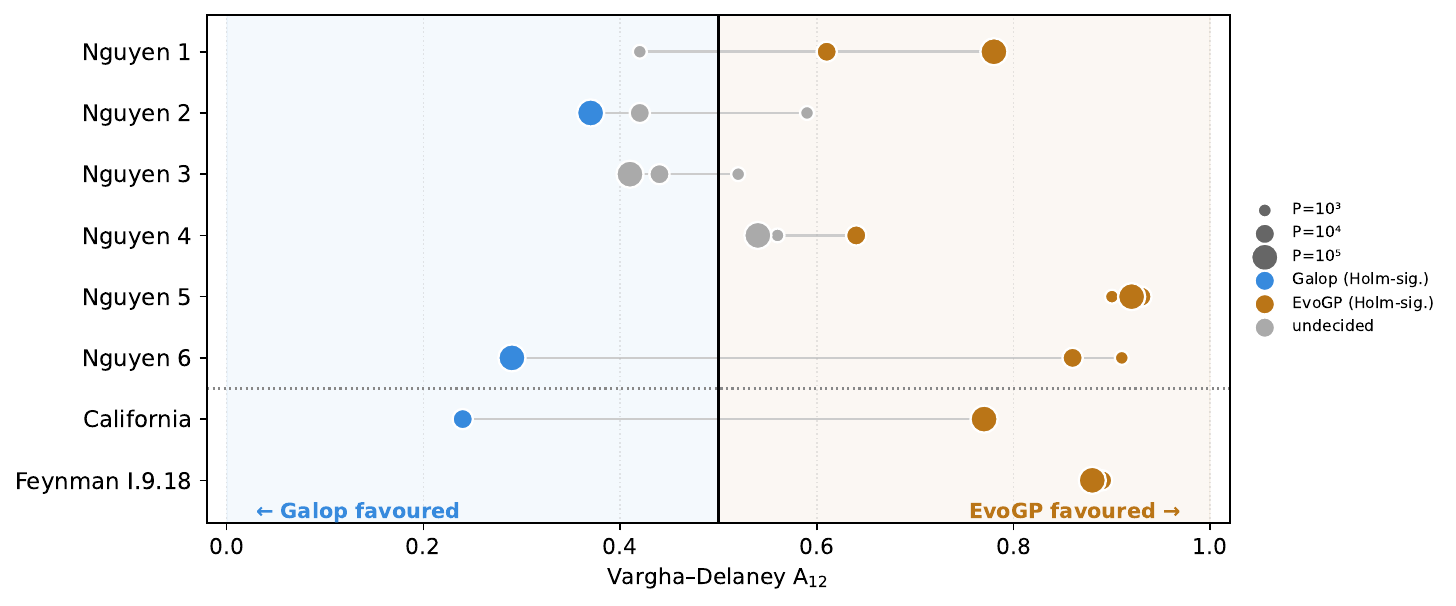}
\caption{
Matched-population quality measured by Vargha--Delaney
$A_{12}$. Dot size denotes population
$10^3$/$10^4$/$10^5$; $A_{12}<0.5$ favours Galop.
Colour indicates the Holm-corrected outcome:
Galop, EvoGP, or undecided.
}
\label{fig:matched-quality}
\end{figure}

Across the 22 matched-population comparisons, the Holm-corrected results show 3 significant wins for Galop, 11 for EvoGP, and 8 undecided outcomes (Fig.~\ref{fig:matched-quality}). EvoGP therefore achieves better quality more often under the tested experimental settings.

On Nguyen~1--6, Galop's significant wins increase from none at $P=10^3$ to two at $P=10^5$. At the largest population, its median test MSE falls below the $10^{-12}$ solve threshold on four of the six problems. However, Nguyen~5 and Feynman favour EvoGP at every tested population, while California Housing favours different systems at the two tested population sizes.

Solve rates provide additional insight beyond error rankings. On Nguyen~1 at $P=10^4$, both systems solve 99\% of runs. EvoGP's advantage reflects small differences among already solved runs and disappears when errors below the solve threshold are treated as equivalent. On Nguyen~4, Galop solves more runs than EvoGP: 28\% versus 7\% at $P=10^4$, and 44\% versus 12\% at $P=10^5$. Nevertheless, the rank-based comparison favours EvoGP at $P=10^4$ and is undecided at $P=10^5$. Thus, error rankings and solve rates capture different aspects of search quality.
An additional comparison using EvoGP's published example settings at $P=10^4$ preserves all EvoGP-favoured outcomes. The two undecided Nguyen comparisons remain undecided, while Galop's California Housing win becomes undecided.

Overall, EvoGP achieves better solution quality more often, although Galop remains competitive on several problems. Sotto et al.~\cite{sotto_graph_2021} similarly found that tree-based GP generally outperformed graph-based and LGP approaches on synthetic symbolic regression benchmarks, while results were more mixed on real-world regression problems.

\subsection{Population Capacity}
\label{sec:results-capacity}

\begin{figure}[!htbp]
\centering
\includegraphics[width=0.95\linewidth]{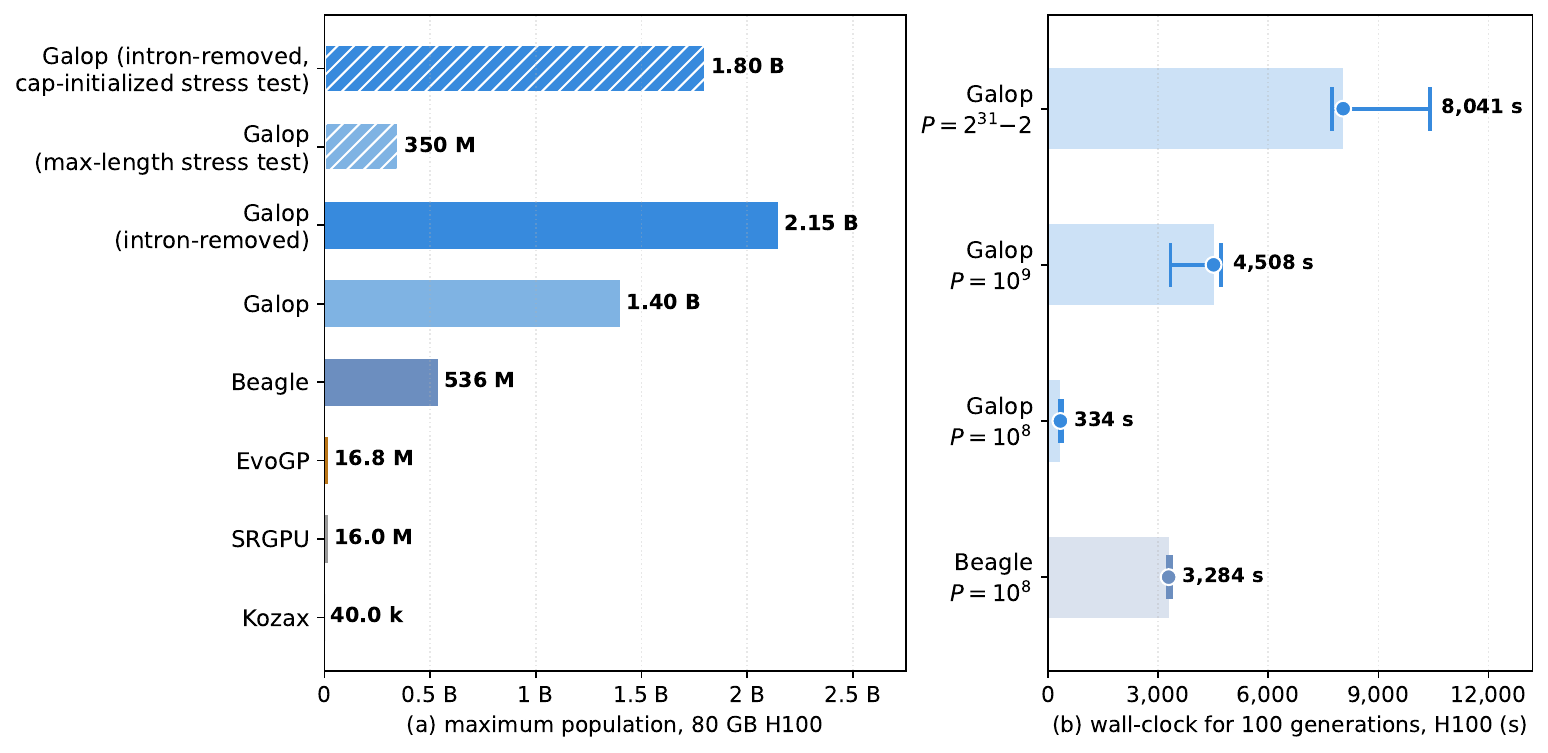}
\caption{
Population capacity on one 80\,GB H100.
(a) Maximum population completing one full generation under the Nguyen~2
capacity protocol; hatched bars show Galop with programs initialized
at the 50-instruction limit.
(b) Wall-clock time for complete 100-generation runs, with intron removal
enabled for all Galop runs. Markers show medians and error bars show
the observed minimum--maximum.
}
\label{fig:capacity}
\end{figure}

Galop supports the largest population measured for a complete generation. Its intron-removed layout completes one generation with $2^{31}-2=2{,}147{,}483{,}646$ individuals, approximately $4.0\times$ the next-largest measured population and $128\times$ EvoGP's (Fig.~\ref{fig:capacity}a). This limit is imposed by Galop's signed 32-bit population index rather than device memory. Without intron removal, Galop reaches $1.4\times10^9$ individuals before device memory becomes limiting.

Using the intron-removed layout, Galop can also sustain these populations across evolution. At $P=10^8$, Galop completes 100 generations in a median 333.8\,s, compared with 3,284\,s for Beagle, a $9.8\times$ difference (Fig.~\ref{fig:capacity}b). All five Galop runs at $P=10^9$ complete 100 generations in a median 4,508\,s, while all five runs at $P=2^{31}-2$ complete in a median 8,041\,s (2.23\,h). 

These maximum population sizes are not normalized for program size, as the systems use different representations and initialization distributions. Galop initializes programs with 8 instructions, compared with an average 3.48 commands for Beagle and 16.3 active nodes for EvoGP. EvoGP also reserves device memory for the configured maximum tree size rather than only its active nodes. To assess Galop under a larger program footprint, we repeat the capacity experiment with every program initialized at its 50-instruction limit. Without intron removal, Galop reaches $3.5\times10^8$ individuals, while enabling intron removal increases this to $1.8\times10^9$ under the standard 16-thread protocol before larger populations exceed the 20-minute runtime limit. This stress test is conservative for Galop rather than a size-normalized comparison between systems.

Population initialization also becomes a substantial cost at these scales. Galop's current population initializer is serial, requiring approximately 400--540\,s at $P=10^9$ and 930--1,300\,s at $P=2^{31}-2$ across the measured nodes. All reported capacity and sustained-run timings include initialization.

Even with intron removal, the maximum population that fits in GPU memory depends on effective program length. Across the measured workloads at $P=10^5$, the mean effective program length (median across runs) after 100 generations ranges from 3.6 to 14.7 instructions. Each individual in Galop's intron-removed layout requires approximately 13 fixed device bytes for offsets, length, and fitness, plus 5 bytes per effective instruction, including allocation headroom. Device memory therefore scales as approximately $13+5L_{\mathrm{eff}}$ bytes per individual, where $L_{\mathrm{eff}}$ is the mean effective program length. This gives estimated population limits near $1.3\times10^9$ for 10 effective instructions and $10^9$ for California Housing's 14.7 instructions. These estimates extrapolate from populations of $10^5$; effective program length may change at larger scales.

Together, these results demonstrate sustained billion-individual evolution on a single GPU. The maximum practical population size depends on program representation, effective code length, and host-side costs.

\subsection{Limitations and Future Work}
\label{sec:future}

\paragraph{Limitations.} The experiments prioritize computational scaling rather than search optimization. Both systems use fixed experimental configurations without problem-specific tuning of mutation, crossover, or other evolutionary hyperparameters. The shared eight-primitive set also restricts the functions available to evolved programs. Quality results should therefore be interpreted as comparisons under the tested configurations rather than estimates of the best performance attainable by either system.

Our billion-scale experiments demonstrate that populations of this size can be evolved for 100 generations. Whether they improve evolutionary search, and on which problems, remains an open question. It is also unclear whether techniques such as diversity maintenance are needed to exploit them, or how they behave over longer runs. Galop makes these questions practical to investigate.

\paragraph{Future work.}
A natural next step is to compare block-, warp-, and finer-grained execution within Galop across different combinations of $P$ and $D$. This would help isolate the effect of execution granularity from other system components. We also plan to extend evaluation beyond symbolic regression to agent-based workloads such as reinforcement learning and control, where sequential state transitions can limit parallelism within each individual. These experiments could identify which execution strategies work best for different workloads and inform adaptive selection of execution granularity.

Multi-GPU execution is another natural extension. The present study deliberately characterizes single-GPU capacity, and the resulting maximum population is not an intrinsic limit of the execution model: Galop's population-level decomposition permits individuals to be partitioned across devices with largely independent fitness evaluation.  This could extend the accessible population scale well beyond the billions demonstrated here.

At extreme populations, host-side work becomes a bottleneck on workloads with small fitness sets. Selection and variation account for an increasing share of runtime, motivating overlap with GPU evaluation or moving more evolutionary operations onto the device. Population initialization is also serial and becomes substantial at billion-scale populations, making parallel initialization a natural target for reducing runtime.

\section{Conclusion}
\label{sec:conclusion}

We presented Galop, a hybrid CPU--GPU linear GP system that evaluates one individual per CUDA warp. Across two unique GPUs, Galop completes evolutionary runs faster than EvoGP in 27 of 28 primary configurations, with speedups of up to $13.6\times$. Its advantage is workload dependent: EvoGP becomes faster when small populations are evaluated against very large fitness sets. At matched population sizes, solution quality favours EvoGP more frequently, although Galop remains competitive on several problems.

Galop's principal contribution is making much larger populations computationally practical. On a single H100, it reaches 2.15 billion individuals, approximately $4.0\times$ the next-largest measured population. All five runs at this population complete 100 generations in a median of 2.23 hours. The maximum practical population size depends on effective program length and host-side costs.

These results establish billion-scale genetic programming as a practical experimental regime on a single GPU. Rather than treating population size primarily as a computational constraint, researchers can begin exploring population scales that were previously prohibitive, including their effects on rare evolutionary events, selection dynamics, and population structure.
\begin{credits}
\subsubsection{\ackname}
This work was supported by the Ontario Graduate Scholarship (OGS)
and the Natural Sciences and Engineering Research Council of Canada
(NSERC) Discovery Grants program. Computational resources were
provided by the Digital Research Alliance of Canada.

We thank Dhilan Teeluckdharry for his assistance in testing Galop,
Sergey Mashchenko for his early guidance and discussions on GPU
computing, and Ali Naqvi for his constructive feedback and discussions.
Xavier Hill Roy also thanks his parents and Laura Danier for their
encouragement and support throughout this work.
\end{credits}
\bibliographystyle{splncs04}
\bibliography{references}

\end{document}